\documentclass[letterpaper, 10 pt, conference]{ieeeconf}  

\IEEEoverridecommandlockouts                              
\makeatletter
\let\NAT@parse\undefined
\makeatother

\usepackage{times}
\usepackage{epsfig}
\usepackage{graphicx}
\usepackage{amsmath}
\usepackage{amssymb}
\usepackage{makecell}
\usepackage{tabularx}
\newcolumntype{L}{>{\raggedright\arraybackslash}X}
\newcolumntype{C}{>{\centering\arraybackslash}X}
\usepackage{array, makecell}
\usepackage{xcolor}
\setcellgapes{3pt}
\usepackage{subfigure}
\usepackage{multirow}

\newcommand*{\crosssymbol}{%
    \text{%
      \raisebox{1ex}{%
        \makebox[0pt][l]{%
          \rule[-.2pt]{.75ex}{.4pt}%
        }%
        \makebox[.75ex]{%
          \rule[-1ex]{.4pt}{1.5ex}%
        }%
      }%
    }%
}   

\usepackage[pagebackref=true,breaklinks=true,letterpaper=true,colorlinks,bookmarks=false]{hyperref}

\title{\LARGE \bf
OpenEDS2020: Open Eyes Dataset
}

\author{Cristina Palmero$^{1 \crosssymbol}$, Abhishek Sharma$^{2}$, Karsten Behrendt$^{3}$, Kapil Krishnakumar$^{3}$, \\Oleg V. Komogortsev$^{2,4}$ and Sachin S. Talathi$^{2}$
\thanks{$^{1}$Universitat de Barcelona, Spain}%
\thanks{$^{2}$Facebook Reality Labs, USA}%
\thanks{$^{3}$Facebook, USA}%
\thanks{$^{4}$Texas State University, USA}%
\thanks{$^{\crosssymbol}$This work was done during internship at Facebook Reality Labs}
}

\begin{document}

\maketitle
\thispagestyle{plain}
\pagestyle{plain}

\begin{abstract}
We present the second edition of OpenEDS dataset, OpenEDS2020, a novel dataset of eye-image sequences captured at a frame rate of 100 Hz under controlled illumination, using a virtual-reality head-mounted display mounted with two synchronized eye-facing cameras. The dataset, which is anonymized to remove any personally identifiable information on participants, consists of 80 participants of varied appearance performing several gaze-elicited tasks, and is divided in two subsets: 1) Gaze Prediction Dataset, with up to 66,560 sequences containing 550,400 eye-images and respective gaze vectors, created to foster research in spatio-temporal gaze estimation and prediction approaches; and 2) Eye Segmentation Dataset, consisting of 200 sequences sampled at 5 Hz, with up to 29,500 images, of which 5\% contain a semantic segmentation label, devised to encourage the use of temporal information to propagate labels to contiguous frames.  Baseline experiments have been evaluated on OpenEDS2020, one for each task, with average angular error of 5.37 degrees when performing gaze prediction on 1 to 5 frames into the future, and a mean intersection over union score of 84.1\% for semantic segmentation. As its predecessor, OpenEDS dataset, we anticipate that this new dataset will continue creating opportunities to researchers in eye tracking, machine learning and computer vision communities, to advance the state of the art for virtual reality applications. The dataset is available for download upon request at \url{http://research.fb.com/programs/openeds-2020-challenge/}. 
\end{abstract}

\section{Introduction}
Eye tracking has emerged as a powerful tool for a number of applications, including health assessment, disease diagnosis~\cite{chita2016social, o2008smooth} and human behavior~\cite{pan2004determinants} and communication~\cite{fan2019understanding} analysis. Nonetheless, the fields that have recently boosted its potential are virtual reality (VR) and augmented reality (AR). Indeed, the potential applications of AR/VR technology to a multitude of sectors such as online education~\cite{fernandez2017augmented}, healthcare~\cite{izard2018virtual, li2017application}, entertainment~\cite{hartmann2020entertainment, pucihar2015exploring}, communication~\cite{smith2018communication, kim2014improving} and/or gaming industry~\cite{thomas2012survey, miller2014effectiveness} have created an ever-growing demand of more realistic and immersive AR/VR experiences.

One of the core technologies that allows to have high quality immersion experience in VR/AR while keeping computational cost of generating the environment low is a technique called Foveated Rending (FR)~\cite{patney2016towards}. FR presents a high-quality picture at the point where a user is looking, while reducing the quality of the picture in the periphery according to a function of human visual acuity. This non-uniform image degradation substantially reduces graphical pipeline’s power consumption without decreasing the perceptual quality of the generated picture. However, fast eye movements present a challenge for FR due to the transmission and processing delays present in the eye tracking and graphical pipelines. If the delays are not compensated for, fast eye movements can take a person’s gaze to the areas of the image that are rendered with low quality, thus degrading the user’s experience. Among the ways of remedying this issue are: a reduction of delays, which is not always possible; predicting future gaze locations, thus compensating for the delays; or a combination of both.

Considering real-time requirements of FR and its goal of reducing power consumption, the prediction of future gaze points based on a short subsequence of the already-estimated gaze locations is considered the most fruitful path of exploration. Accurate eye movement prediction requires high quality underlying signal to be effective. One of the important metrics describing the quality of the captured raw eye positional signal is spatial accuracy which can be linked to the accuracy of eye segmentation when machine learning approaches are employed for gaze estimation. It must be noted that spatial accuracy is critical not only for eye movement prediction, but also for applications such as health assessment, and direct gaze interaction.

The two predominant approaches for image/video-based gaze estimation, also known as \emph{video occulography}, can be broadly classified into geometric and appearance-based systems~\cite{hansen2009eye}. Geometric approaches treat the human eye as a sphere-on-sphere model \cite{guestrin2006general} and find the pupil and glint locations in a precisely calibrated system of cameras and LEDs to infer the 3D location of the pupil and the gaze direction. More recent geometric approaches do not require dedicated systems and/or glints, and make use of 3D morphable models instead~\cite{wang2017real,wood20163d}. On the other hand, appearance-based models are typically based on end-to-end inference models, such as deep Convolutional Neural Networks (CNNs) \cite{zhang2017mpiigaze, park2018deep}, to directly estimate the gaze direction in the frame of reference of the camera. Nowadays, both geometric and appearance-based approaches usually rely on one or more deep learning modules to tackle the large variations in eye appearances due to anatomical differences, lighting, camera viewpoint and/or makeup across the human population. While the current state-of-the-art appearance-based methods exploit end-to-end deep networks to directly regress gaze from input eye/face images, geometric methods deploy highly accurate segmentation networks to extract pupil, iris, sclera and skin regions for further processing~\cite{yiu2019deepvog}. 

In real-world applications, the input to any gaze estimation system are temporal sequences of eye-images in the form of a video. However, most popular approaches do not exploit temporal information and instead estimate gaze direction for each frame separately. Intuitively, there exists useful temporal information in videos that can be leveraged to improve the current gaze estimation approaches. For example, incorporating the velocity and direction of the eyeball motion can be leveraged to improve upon gaze estimation~\cite{palmero2020toappear} and semantic segmentation accuracy. While a number of remote-camera-based works have recently started to exploit this area of research~\cite{palmerorecurrent, wang2019neuro}, the lack of high-resolution, real-world datasets with images sampled at a sufficient sampling rate to capture small and fast eye movements, such as saccades, with accurate ground-truth annotations, is affecting further progress on the topic.

In this paper, and motivated from the aforementioned gap in the literature, we release OpenEDS2020, a large-scale anonymized dataset of sequences of high-resolution eye-images (640 $\times$ 400 pixels) sampled at 100 Hz with semantic segmentation and gaze direction annotations. This dataset provides higher resolution and higher frame rate image sequences in comparison to existing publicly available datasets, as well as accurate gaze direction and semantic segmentation annotations. We believe OpenEDS2020 can be instrumental to advance the current state of the art in gaze estimation research and benchmarking of existing algorithms. 

\section{Data collection}

\begin{figure*}
	\centering
	\subfigure{\includegraphics[width=4cm]{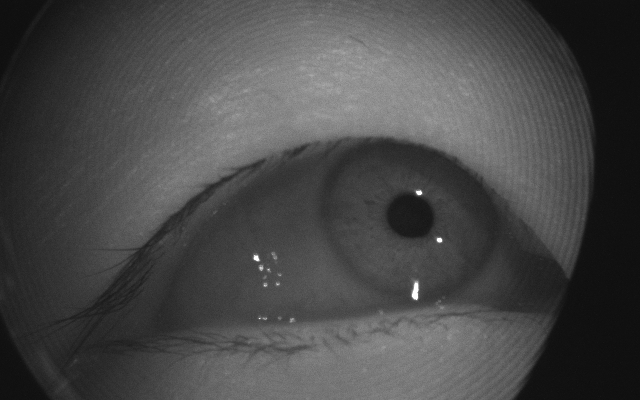}}
	\hfill
	\subfigure{\includegraphics[width=4cm]{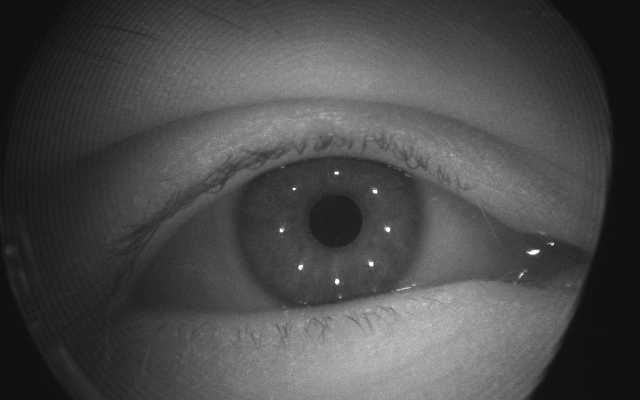}}
	\hfill
	\subfigure{\includegraphics[width=4cm]{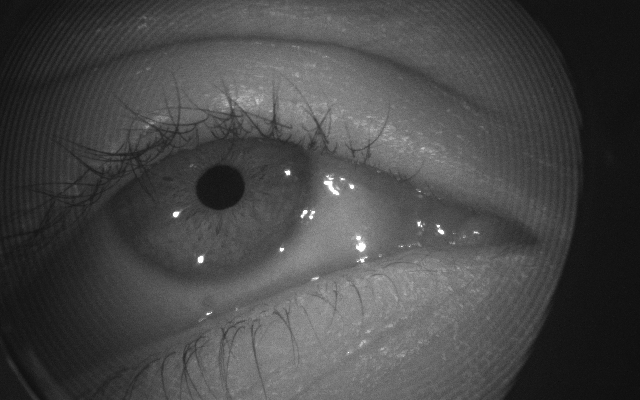}}
	\hfill
	\subfigure{\includegraphics[width=4cm]{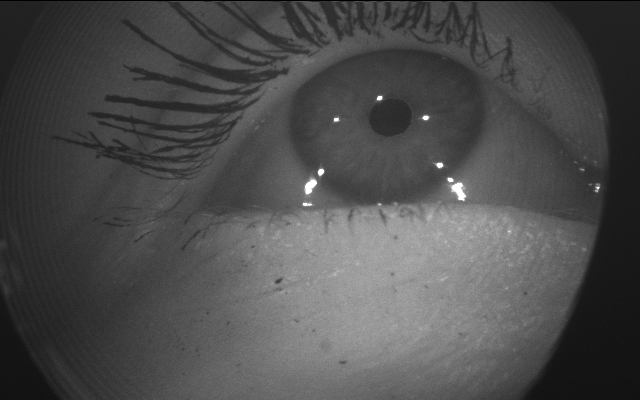}}\\
	\subfigure{\includegraphics[width=4cm]{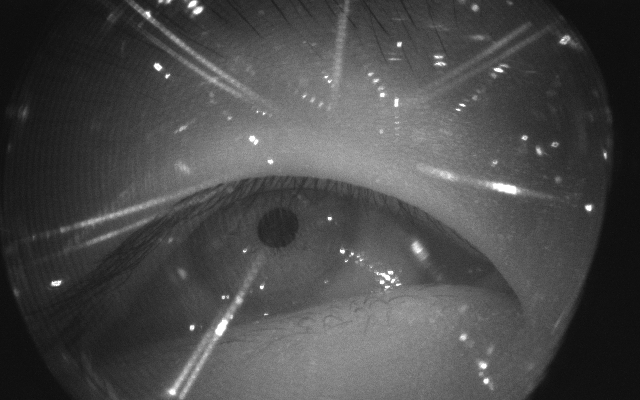}}
	\hfill
	\subfigure{\includegraphics[width=4cm]{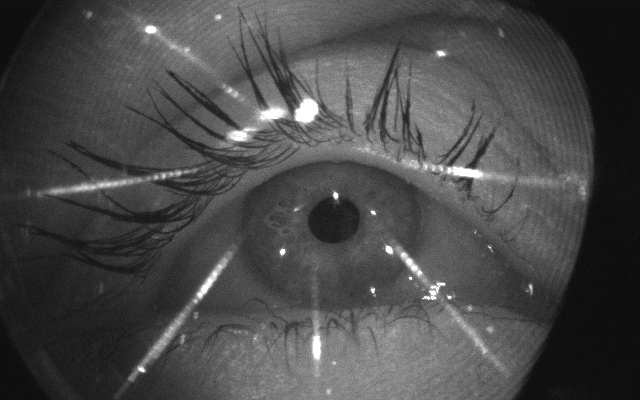}}
	\hfill
	\subfigure{\includegraphics[width=4cm]{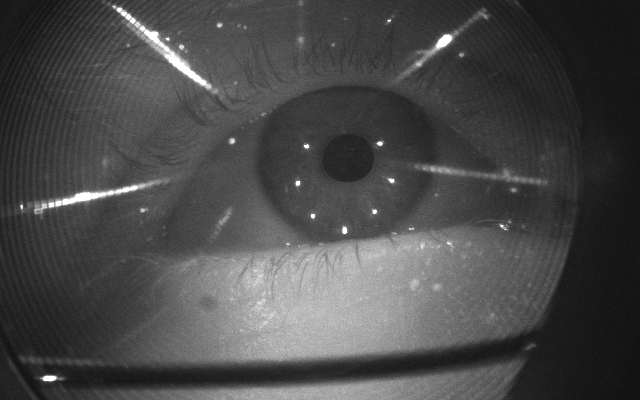}}
	\hfill
	\subfigure{\includegraphics[width=4cm]{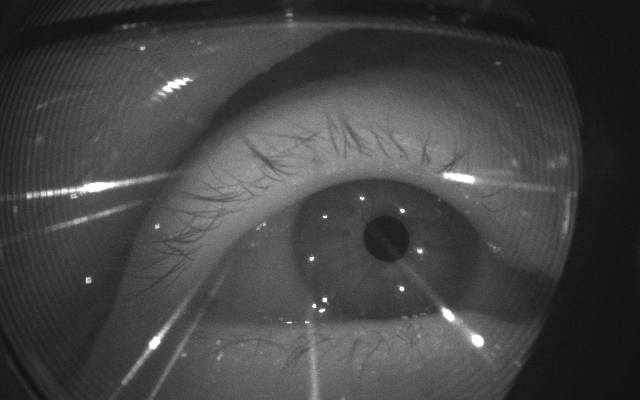}}
	\caption{Examples of images without glasses (top row) and with glasses (bottom row), representing the variability of the dataset in terms of accessories, ethnicity, age and gender.}
	\label{fig:image_examples}
\end{figure*}

The dataset was captured from 90 voluntary participants of ages between 20 and 70 with appearance variability in terms of ethnicity, gender, and age, and some of them wearing glasses, contact lenses, and/or make-up. These participants provided written informed consent for using their eye images for research and commercial purposes before taking part in the data collection stage. Participants were asked to wear a VR head-mounted display (VR-HMD), mounted with two synchronized eye-facing infrared cameras at a frame rate of 100 Hz, and were recorded while gazing at specific dot-patterns displayed on a blank screen with different target movements. The dataset was anonymized to remove any personally identifiable information on the participants for further processing, as explained in Sections 3 and 4.

Each recording consisted of a set of patterns as follows: ring-shaped patterns at $\pm$20 degrees eliciting smooth pursuit movements, and random point changes to induce saccades, where the targets were moved in a $\pm$20 degree cone from 50 cm in depth to 600 cm. Other variations included changing the brightness of the background from light to dark to ensure the effects of pupil dilation were captured, and moving the headset while recording to simulate slippage. Each recording session lasted approximately 5 minutes. Example images of the collected dataset are shown in Figure \ref{fig:image_examples}.

\section{Gaze prediction}

\begin{figure*}
	\centering
	\subfigure{\includegraphics[width=0.09\linewidth]{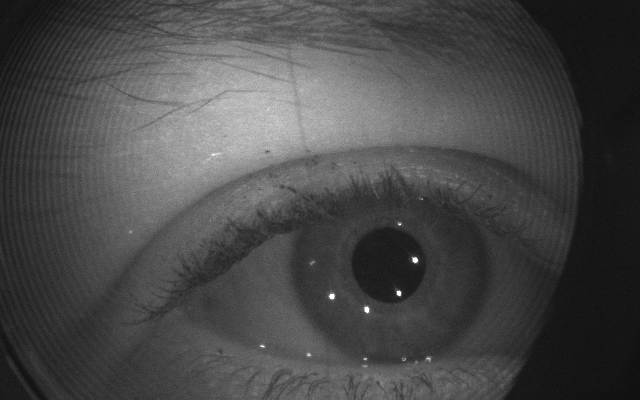}}
	\subfigure{\includegraphics[width=0.09\linewidth]{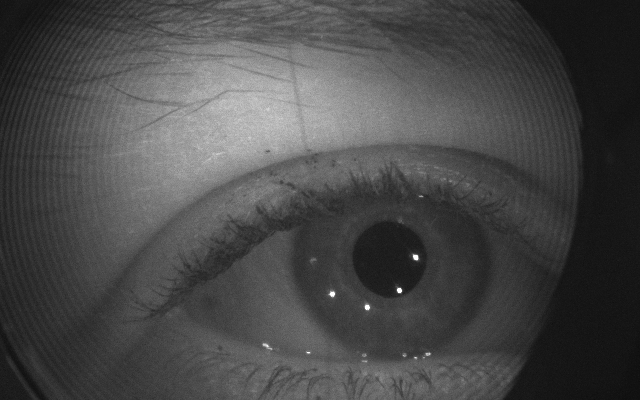}}
	\subfigure{\includegraphics[width=0.09\linewidth]{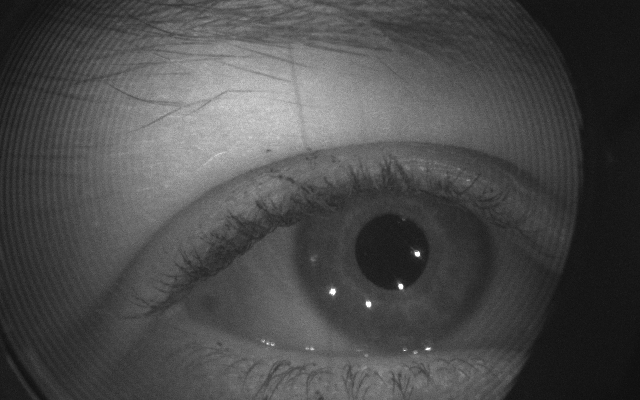}}
	\subfigure{\includegraphics[width=0.09\linewidth]{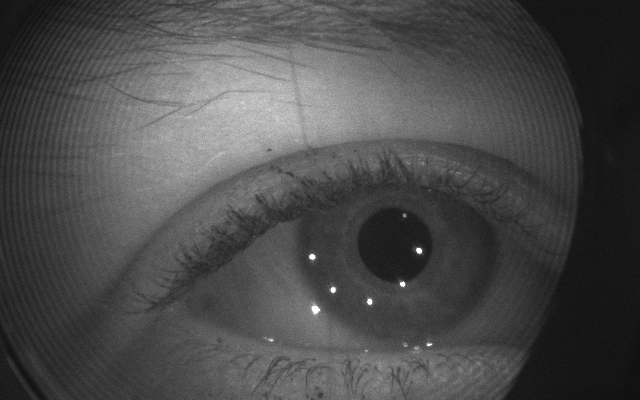}}
	\subfigure{\includegraphics[width=0.09\linewidth]{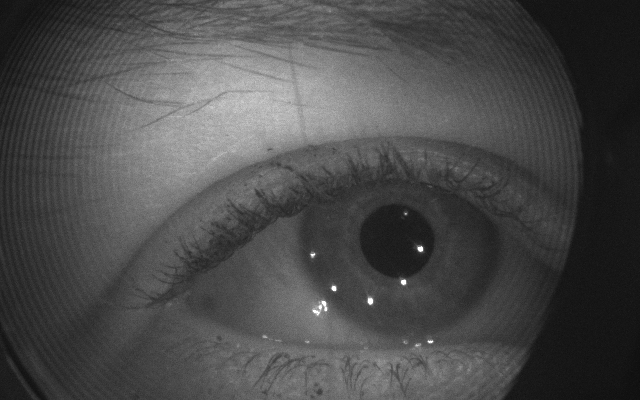}}
	\subfigure{\includegraphics[width=0.09\linewidth]{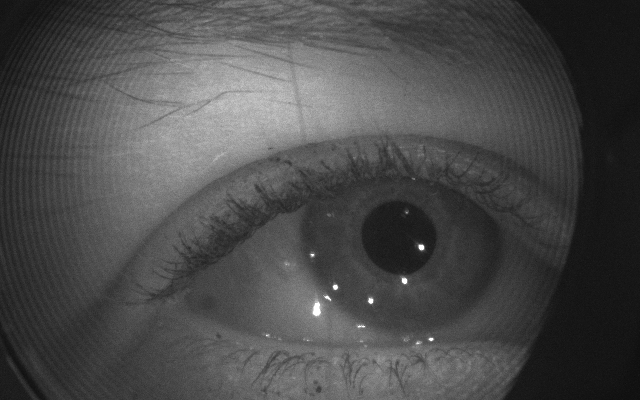}}
	\subfigure{\includegraphics[width=0.09\linewidth]{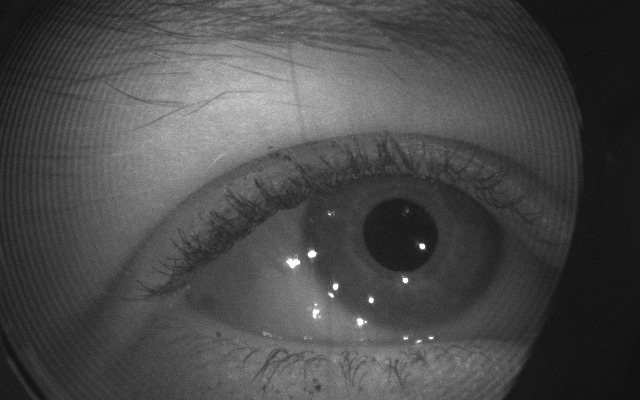}}
	\subfigure{\includegraphics[width=0.09\linewidth]{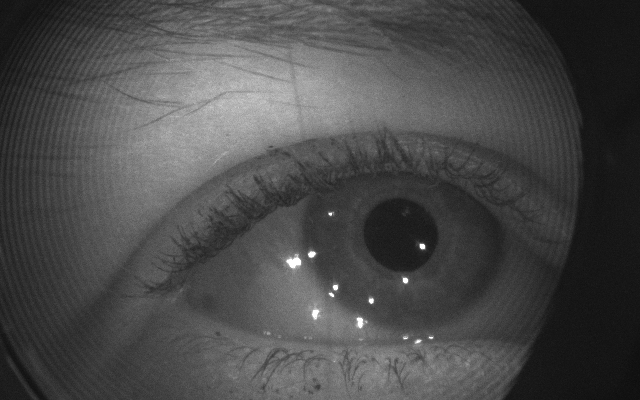}}
	\subfigure{\includegraphics[width=0.09\linewidth]{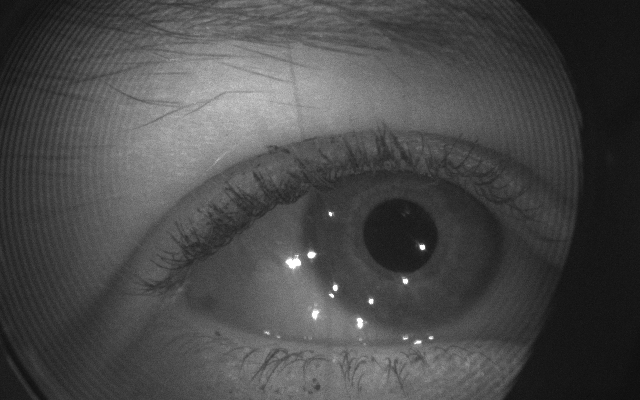}}
	\subfigure{\includegraphics[width=0.09\linewidth]{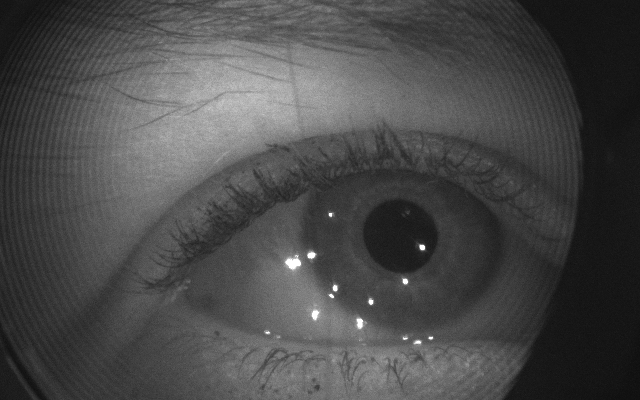}} \\
	\vspace{-0.5em}
	\subfigure{\includegraphics[width=0.09\linewidth]{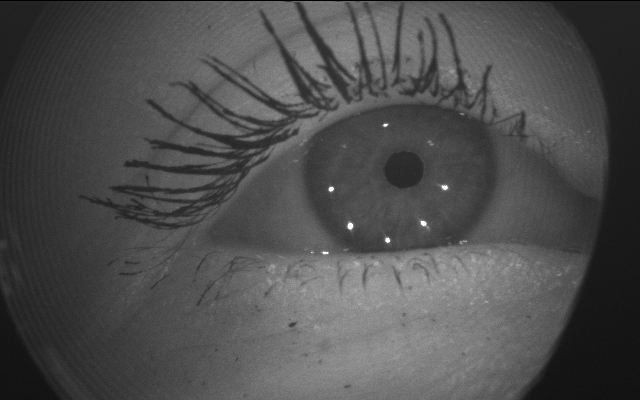}}
	\subfigure{\includegraphics[width=0.09\linewidth]{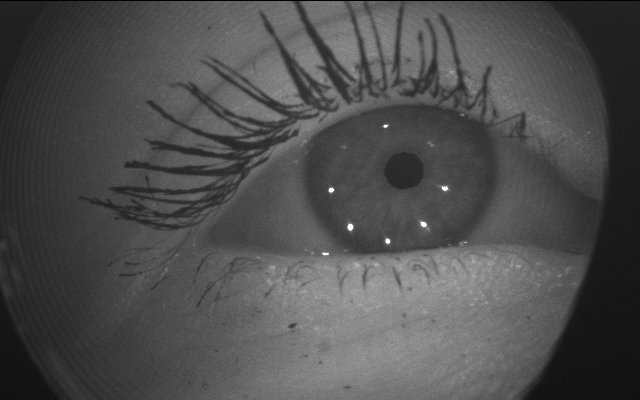}}
	\subfigure{\includegraphics[width=0.09\linewidth]{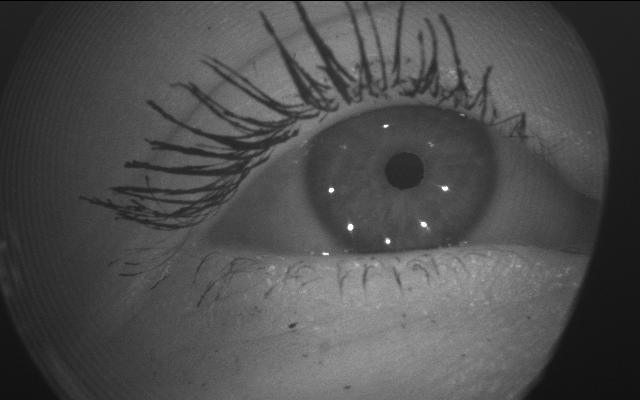}}
	\subfigure{\includegraphics[width=0.09\linewidth]{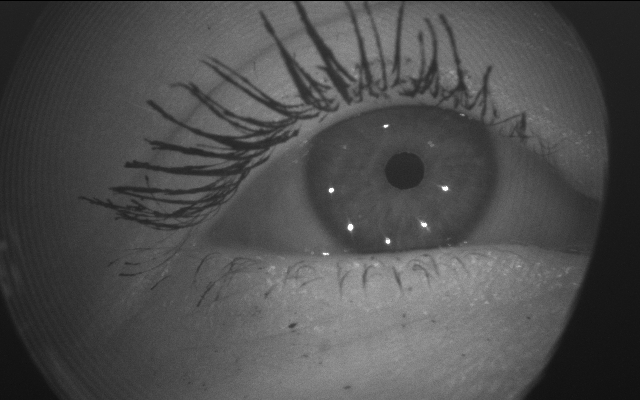}}
	\subfigure{\includegraphics[width=0.09\linewidth]{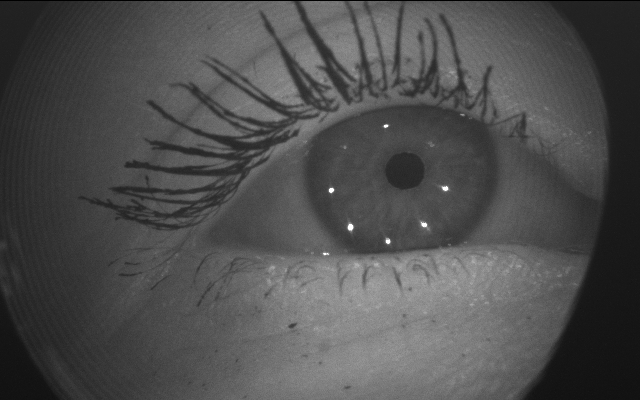}}
	\subfigure{\includegraphics[width=0.09\linewidth]{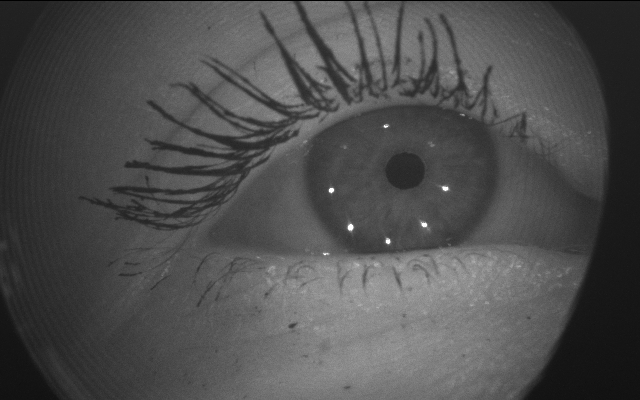}}
	\subfigure{\includegraphics[width=0.09\linewidth]{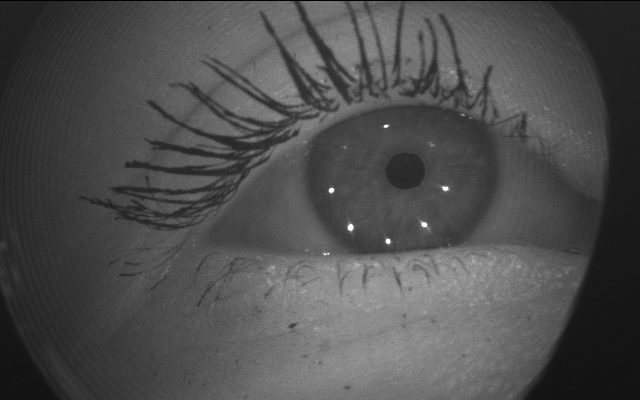}}
	\subfigure{\includegraphics[width=0.09\linewidth]{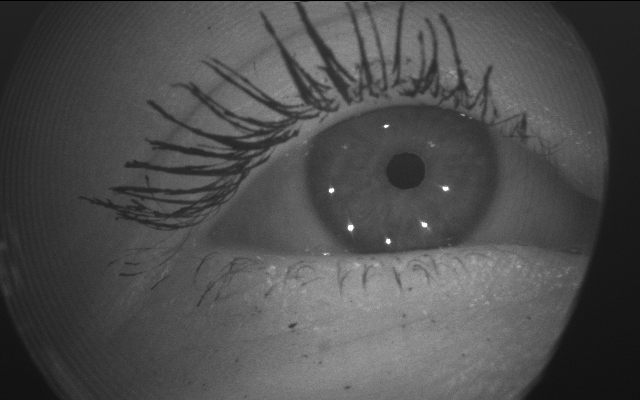}}
	\subfigure{\includegraphics[width=0.09\linewidth]{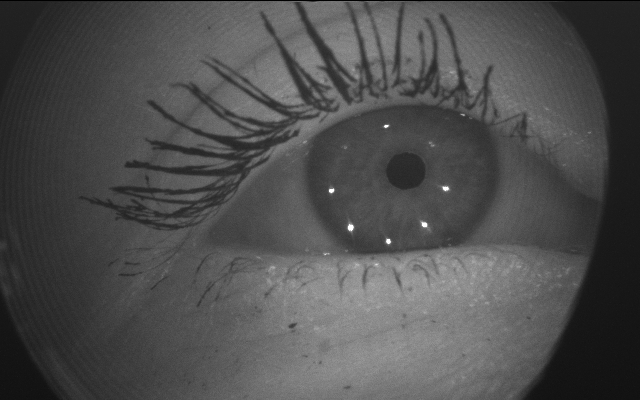}}
	\subfigure{\includegraphics[width=0.09\linewidth]{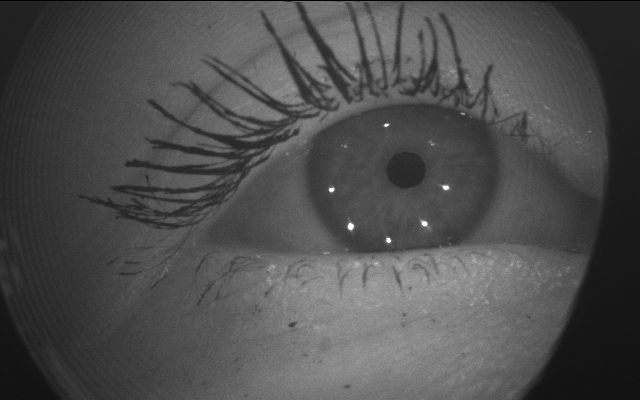}}
	\caption{Example of saccadic (top row) and smooth pursuit (bottom row) eye movements during 100ms.}
	\label{fig:image_examples_movements}
\end{figure*}

This subset of data is aimed at fostering research on spatio-temporal methods for gaze estimation and prediction for tasks involving different eye movements, such as saccades, fixations and smooth pursuit. Examples of different eye movements are illustrated in Fig. \ref{fig:image_examples_movements}. Based on the eye movement characteristics~\cite{purves2001types} and the frame rate of our dataset, we propose to predict 1 to 5 frames (10 to 50 ms) into the future, which is a useful range in AR/VR applications~\cite{albert2017latency}. Furthermore, we hypothesize that 50 frames (500 ms) is a reasonable number to set as the maximum amount of frames that can be used to initialize a gaze prediction model, and design the dataset based on these decisions.  

\subsection{Dataset curation}
Two types of patterns were selected for this data subset, due to the differences in eye movement dynamics associated to them: 
\begin{itemize}
    \item Smooth pursuit-elicited pattern, intertwining 1s fixations at fixed depths with 1s-long smooth pursuit movements as smooth transitions between fixation points, with a total of 17 fixations and 16 transitions per recording.
    \item Saccade-elicited pattern, with 1s-long randomly-located target fixations at different depths and instantaneous (0.1s) target transitions, with up to 20 fixations per recording. 
\end{itemize}

All subjects were recorded following both pattern types and the collected data was anonymized. Frames with invalid ground truth gaze vector, due to either subject distractions, blinks or incorrect estimates (see Section \ref{sec:pred_anot}), were manually discarded, maintaining 80 out of the initial 90 subjects. The remaining data was further processed by randomly selecting $s$ non-overlapping sequences of $f$ contiguous frames each per pattern, with a maximum of 100 frames (1s) per sequence. This way, each sequence can contain either fixations only, smooth pursuit movement only, combination of fixation and smooth pursuit, and combination of fixation and saccadic movement. Each eye was processed individually, and right eyes (from the camera point of view) and respective ground truth vectors were flipped horizontally to seamlessly augment the data. 

We believe that an eye tracking dataset designed for potential spatio-temporal methods should contain a sufficiently representative gaze angle distribution and appearance variability to train a gaze estimation model, while ensuring variability in terms of eye movements, directions and velocities to train a gaze prediction model. We therefore devise three different subject-independent partitions of our dataset, with 32 subjects for training, 8 for validation and 40 for test. To do so, we performed stratified sampling in terms of gender, ethnicity, age and glasses, to ensure having a representative sample for testing. 

\begin{figure*}
	\centering
	\subfigure{\includegraphics[width=5cm]{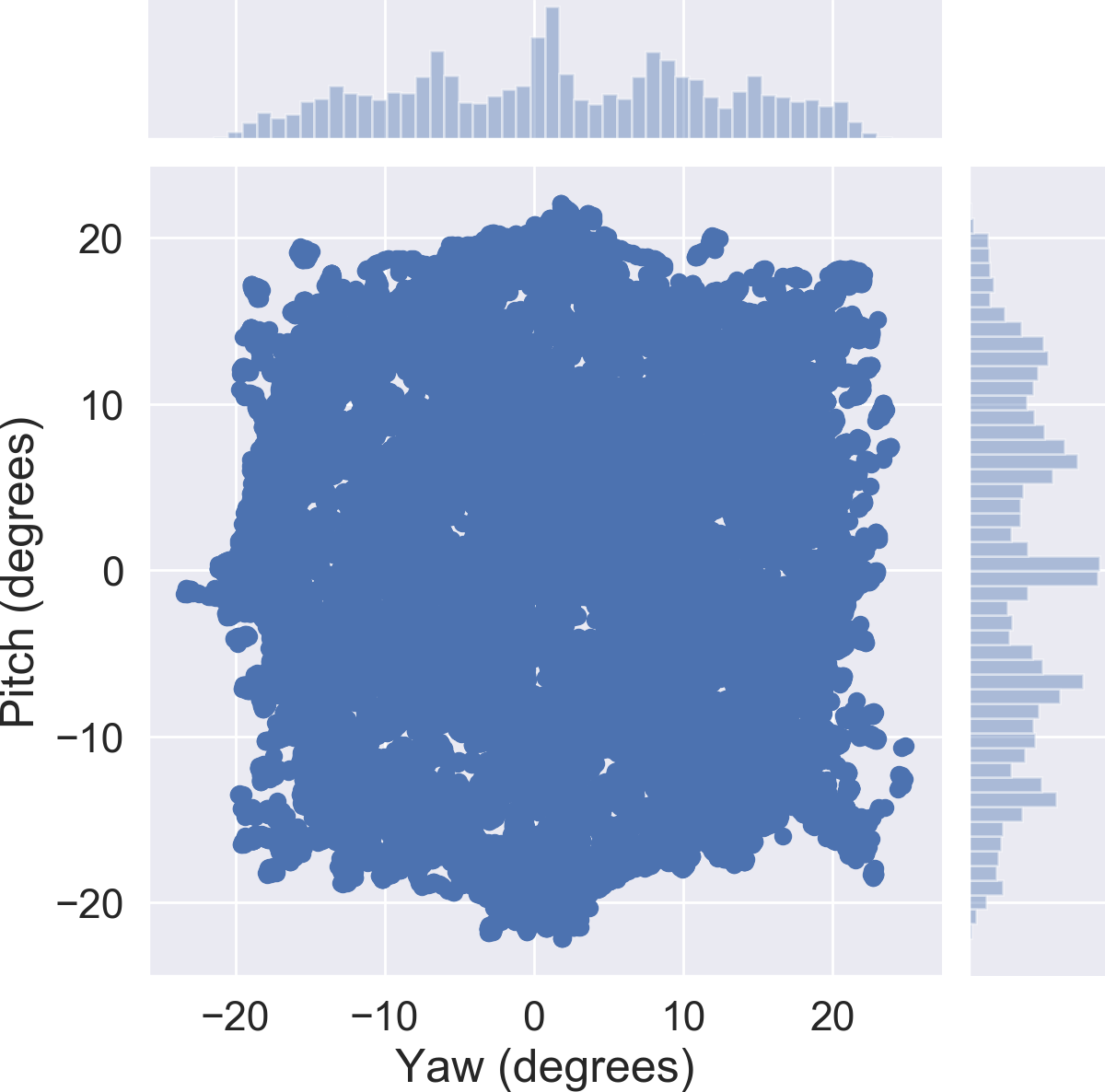}}
	\hfill
	\subfigure{\includegraphics[width=5cm]{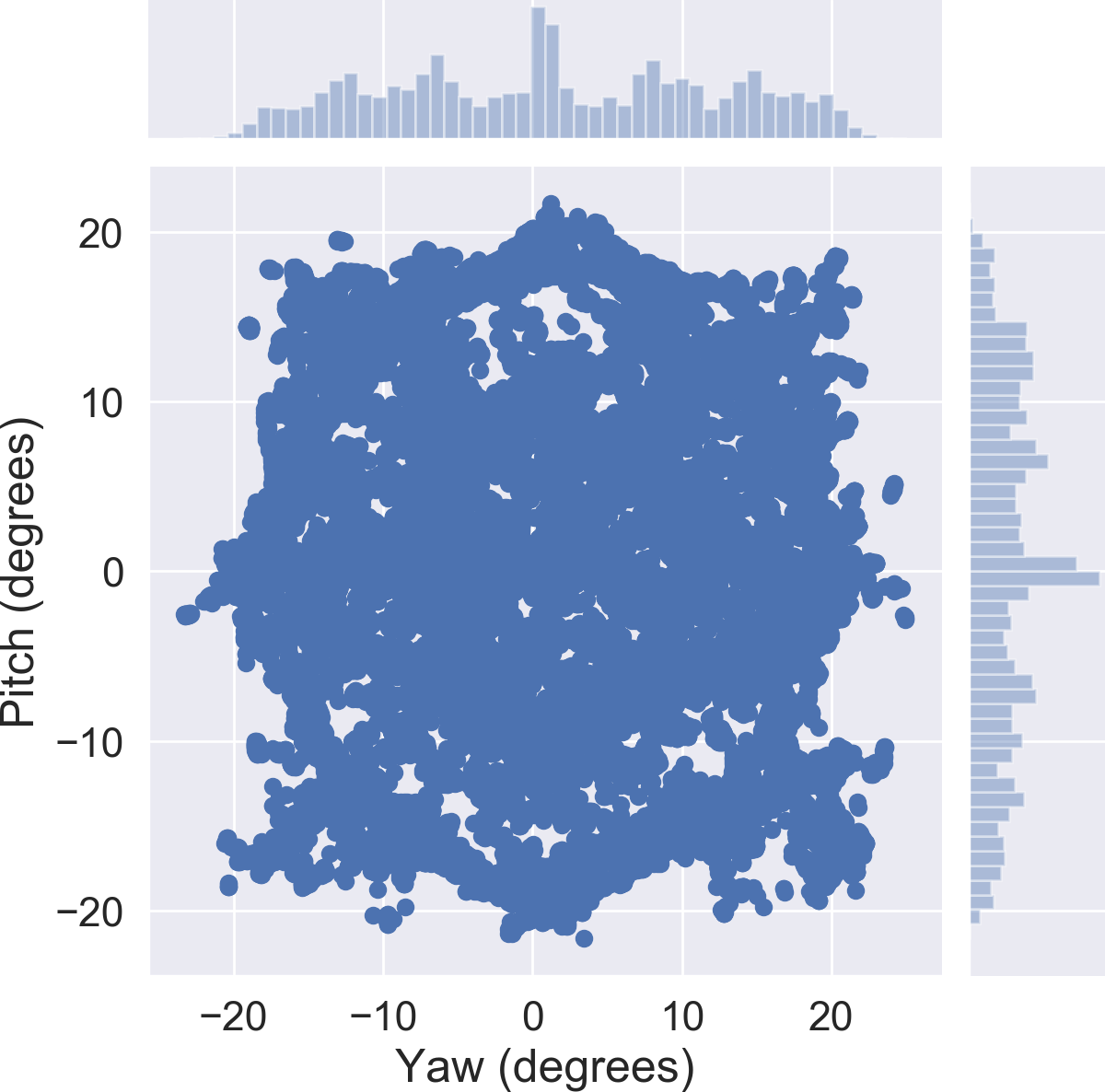}}
	\hfill
	\subfigure{\includegraphics[width=5cm]{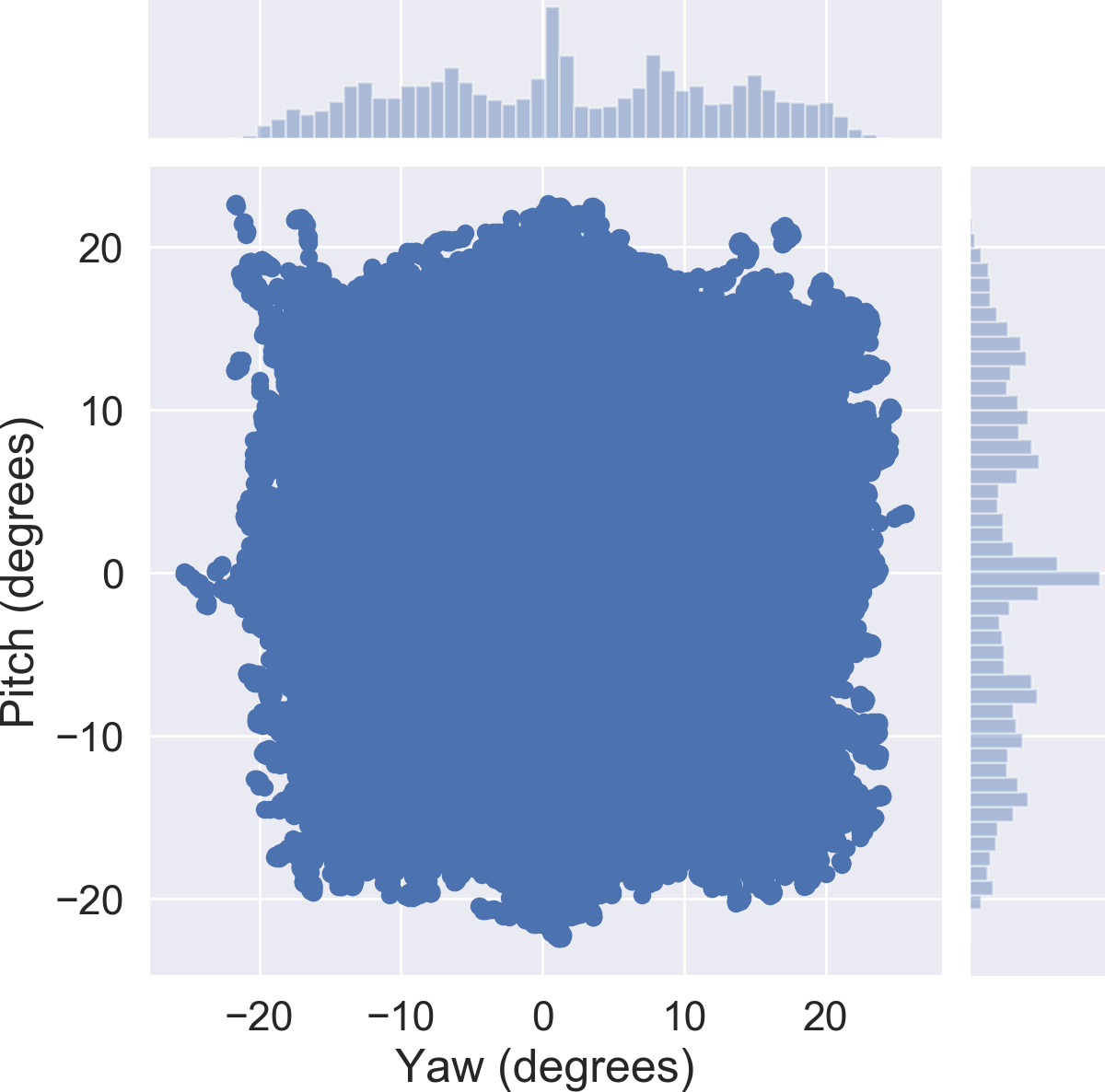}}
	\caption{2D gaze angle distributions for train (left), validation (center) and test (right) splits of gaze prediction data subset.}
	\label{fig:distribution}
\end{figure*}

The training subset consists of 10 100-frame sequences per eye, type of pattern and anonymized subject, with a total of 4,000 images per subject and 128,000 images in total. Assuming that we can use up to 50 frames to initialize a gaze prediction model to predict up to 5 frames into the future, using a sliding window of stride 1 allows us to obtain up to 46 subsequences in a 100-frame sequence, which sums up to 58,880 final sequences in the training set (i.e. 1,280 100-frame sequences in which a sliding window of stride 1 and size 55 is used to create subsequences). For validation and test subsets, however, we chose to use 55-frame sequences so as to have one set of initialization frames and predictions per sequence, which facilitates evaluation and analysis. To compensate for the difference in number of effective sequences with respect to the training set, we selected approximately 80 55-frame sequences per type of pattern and subject, with 70,400 images for validation and 352,000 for test. For subjects for which there was not enough valid data to obtain such 80 sequences per pattern, the maximum amount of valid sequences was selected, and the remaining sequences were obtained from other subjects until we obtained the desired number of sequences, 1,280 and 6,400, respectively. Note that the goal of training and validation subsets is both gaze estimation and prediction, while the goal of the test subset is gaze prediction. Therefore, even though the number of images is substantially bigger for the test set, the number of sequences is what we focus on. 
The distribution of ground truth gaze angles (horizontal and vertical component of gaze) for each data split is depicted in Fig. \ref{fig:distribution}, and their general statistics in terms of subjects variability and number of images and sequences per split shown in Table \ref{tab:stats_subject}.

\subsection{Annotations} \label{sec:pred_anot}
Ground truth 3D gaze vectors are provided per each eye image. Gaze vectors and cornea centers of each eye were obtained using a hybrid model, which combines a deep eye segmentation model (see Section \ref{sec:seg_method}) with a classical user-calibrated glint-based model~\cite{guestrin2006general}. Since these models are frame-based and thus may produce fluctuating estimates, a median filter of window size 5 was used to temporally smooth resulting gaze vectors. An offset correction was further applied to them per subject and pattern, by subtracting the average difference between cornea-to-target vectors and gaze vectors.

\begin{table*}[t!]
\footnotesize
	\begin{tabular}{| c | c | c | c | c |c |c |c |c |c |c |c |c |c | }
		\hline
		\multirow{2}{*}{} & %
		\multicolumn{2}{c|}{Gender} & %
		\multicolumn{3}{c|}{Ethnicity} & %
		\multicolumn{4}{c|}{Age} & %
		\multicolumn{2}{c|}{Accessories} & %
		\multicolumn{2}{c|}{Number of} \\ %
		\cline{2-14}
			  & Female & Male & Asian & Caucasian & Other & 21-25 & 26-30 & 31-40 & 41+ & Glasses & Make-up  & Images & Seqs.     \\ \hline
		Train & 9 & 23 & 10 & 15 & 7 & 6	& 7 & 13 & 6  & 8 & 5& 128k & 1,280 ($\times$ 46)  \\ \hline
		Val.  & 3 & 5 & 1 & 4 & 3 & 2 & 1  & 3 & 2 & 2 &  0 & 70.4k & 1,280   \\ \hline
		Test  & 12 & 28 & 16 & 17 & 7 & 10 & 10 & 14 & 6	& 11 & 5 & 352k & 6,400  \\ \hline
		Total &  24 & 56 & 27 & 36 & 17 & 18 & 18 & 30	& 14 & 21 & 10 & 550.4k & 8,960 (66,560)  \\ \hline 
	\end{tabular}
	\caption{Statistics of the gaze prediction data subset.}
	\label{tab:stats_subject}
\end{table*}

\section{Semantic eye segmentation}\label{subsec:seg}
This subset of data is aimed at exploiting the temporal information available in the form of short temporal sequences of eye images to improve the semantic segmentation accuracy achieved by treating each frame separately. While there could be multiple approaches to exploit temporal information for improving semantic segmentation, we resort to a simple and practically useful problem of accurate label propagation from a few labeled images to all the frames in a sequence. On one hand, this problem can serve as a test-bed and future improvements for the latest deep-learning based spatio-temporal models for real-time inference in videos, few-shot learning, geometry-constrained semantic segmentation and/or co-segmentation. On the other hand, this problem also helps in generating high-quality pseudo ground truth data generation for training semantic segmentation networks.

\begin{table*}[!t]
\footnotesize
    \centering
    \begin{tabular}{|c|c|c|c|c|c|c|c|c|c|c|c|c|c|c|}
    	\hline
         \multicolumn{2}{|c|}{Gender} &
         \multicolumn{3}{c|}{Ethnicity} &
         \multicolumn{4}{c|}{Eye Color} &
         \multicolumn{2}{c|}{Accessories} &
         \multicolumn{4}{c|}{Age}\\
         \hline
         Male & Female & Asian & Caucasian & Other & Brown & Blue & Hazel & Green & Glasses & Make-up & 21-25 & 26-30 & 31-40 & 41+\\
         \hline
         47 & 27 & 31 & 30 & 13 & 50 & 14 & 4 & 6 & 65 & 14 & 17 & 15 & 25 & 16\\
         \hline
         
    \end{tabular}
    \caption{Statistics of the 200 selected sequences for the eye segmentation subset, in terms of gender, ethnicity, eye color and accessories (i.e. glasses, make-up).}
    \label{tab:sem_stat}
\end{table*}

\subsection{Dataset curation}
\begin{figure}[t!]
    \centering
    \includegraphics[width=1\linewidth]{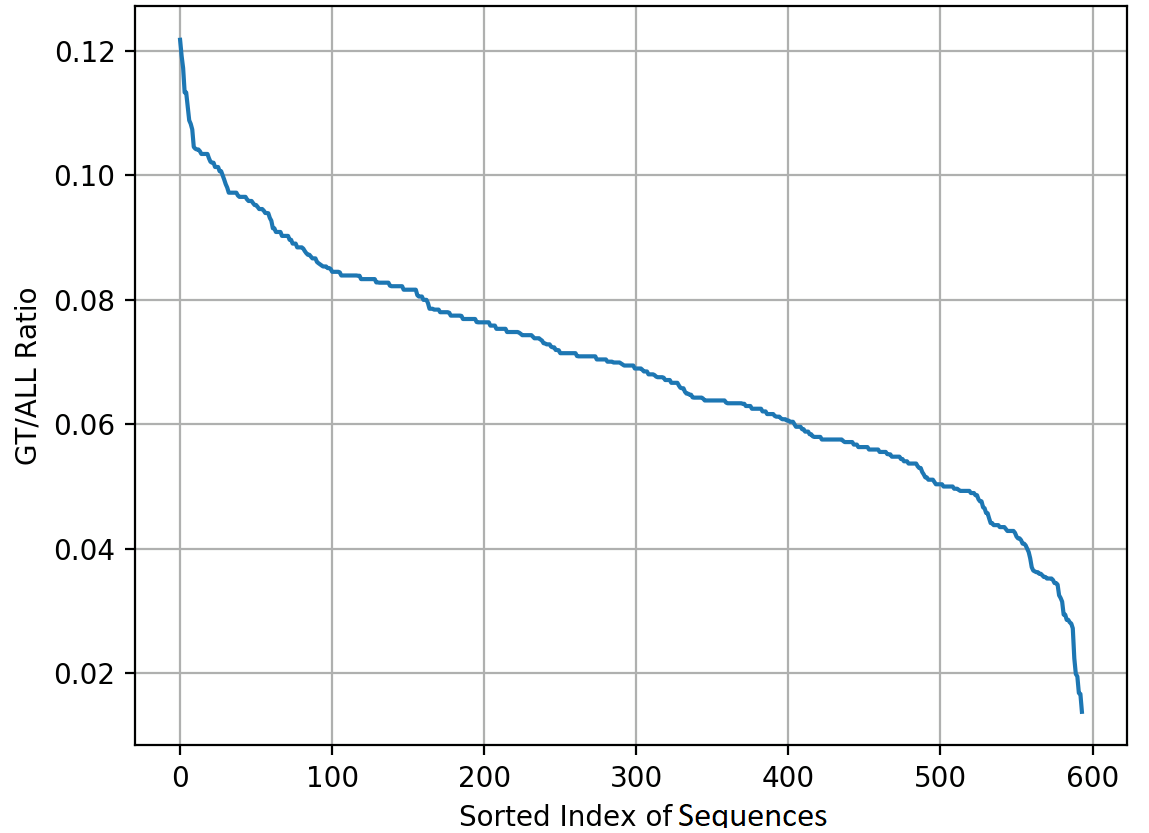}
    \caption{The ratio of labels vs.\ total number of samples for all the 594 sequences shown in the decreasing order.}
    \label{fig:ratio}
\end{figure}

We start the curation process with the initial data, that is, a total of $\sim$600K images in the form of 594 temporal sequences and 11,476 hand-annotated semantic segmentation masks chosen randomly for annotation. Let us define the \emph{label-ratio}, $R$, as: 
\begin{equation}
    R^i = \frac{N_{label}^i}{N_{total}^i} \times 100\%
\end{equation}
here, $N_{label}^i$ and $N_{total}^i$ are the number of labeled samples and the total number of samples in the $i^{th}$ sequence $S_i$, respectively. For this dataset, we decided to set $R\approx5\%$, or in other words, provide $\sim$5\% labeled samples for each sequence. This choice is motivated from the practical considerations of the size of the available dataset and annotations. In order to maintain $R \approx 5\%$ for training, we subsampled the data at 5Hz and chose top 200 sequences sorted in decreasing order of $R$, which resulted in $\sim$150 frames ($\sim$30 seconds of recording) for a total of 29,476 frames with 2,605 semantic-segmentation annotation masks. Fig.~\ref{fig:ratio} shows the ratio of labels vs.\ total number of samples for all the 594 sequences. Out of the total 2,605 annotations, we randomly hide 5 samples per sequence as the test samples, that eventually leads to $R\approx5\%$ for training and 3\% for testing. The 200 sequences were obtained from 74 different subjects. Additional statistics for this dataset can be found in Table~\ref{tab:sem_stat}. Note that the statistics are presented at a complete sequence level and not at frame level. 

\subsection{Annotations}

\begin{figure}[t!]
	\centering
	\subfigure{\includegraphics[width=0.47\linewidth]{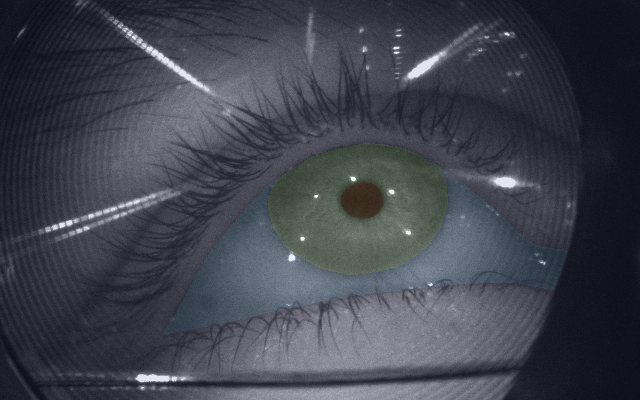}}
	\hfill
	\subfigure{\includegraphics[width=0.47\linewidth]{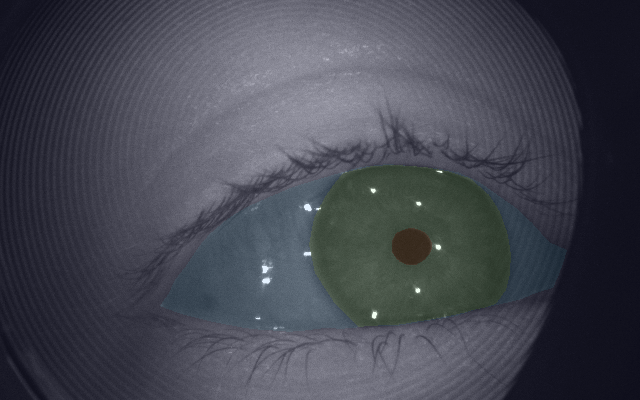}}\\
	\subfigure{\includegraphics[width=0.47\linewidth]{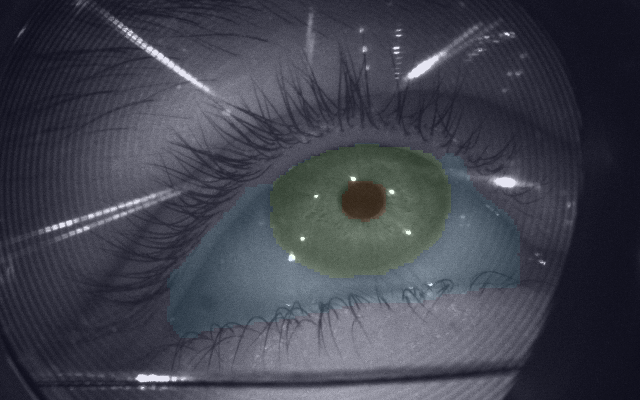}}
	\hfill
	\subfigure{\includegraphics[width=0.47\linewidth]{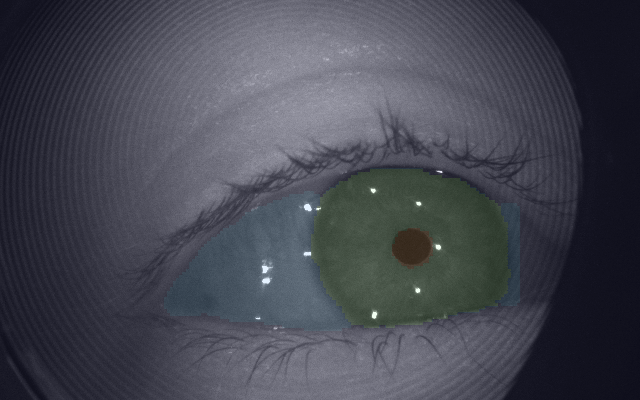}}\\
	\caption{Examples of human annotations (top row) and baseline model performance (bottom row) for eye segmentation.}
	\label{fig:segmentation_examples}
\end{figure}

Human annotations are provided in the form of pixel-level segmentation masks with the following labels: 1) eye region, 2) iris, and 3) pupil. Annotation quality was evaluated by estimating the mean Intersection Over Union (mIOU) score of annotations, ensuring that no labels produced by at least two annotators have less than 80\% mIOU.
The right eye was flipped horizontally to align tear ducts across left and right eye annotations, so the annotators have consistency. The tear ducts were labeled as part of the eye region. In half occlusions, the annotations only include visible parts of the pupil and iris, and eyelashes are labeled as part of the underlying pupil or iris region. Finally, during blinks a thin sliver of the eye region is labeled. Examples of human annotations are provided in Fig. \ref{fig:segmentation_examples}.

\section{Baseline methodologies}
In this section, we describe and evaluate a set of baseline models for each data subset to demonstrate the usefulness and validity of the data for the suggested tasks.

\subsection{Gaze prediction}
We propose a simple baseline model for the Gaze Prediction dataset, in which spatio-temporal information is not jointly leveraged. Instead, we disentangle gaze estimation from prediction, estimating first the line of gaze from each eye image individually, and then performing gaze prediction based on the previously-estimated gaze vectors. We report gaze estimation and prediction results in terms of angular error in degrees between estimated/predicted and ground truth 3D gaze vectors.

For gaze estimation, we train a deep-CNN based on a modified ResNet of 13 convolutional layers, as in \cite{palmero2020toappear}. The CNN backbone was coupled with a 32-hidden unit fully connected layer (FC) and a 2-hidden unit FC linear regression layer, to estimate 2D gaze angles. To compensate for the gaussian-like distribution of the data, we fit a multivariate Gaussian to the training set and weight the samples with their inverse probability. The network was trained end-to-end on 75\% of the training data following such weighting scheme for 50 epochs with ADAM optimizer, initial learning rate of 0.001 and batch size of 32. For training, data was randomly augmented in terms of brightness, horizontal and vertical shifts and rotation. Model training and inference was performed on downsampled images of 160$\times$100 pixels. The total number of parameters of the model was 206k and the resulting model size is 828KB. The trained gaze estimation network is tested on the validation set, obtaining an average error of 4.58 degrees, which is in line with state-of-the-art subject-independent approaches \cite{palmerorecurrent}.

Our gaze prediction approach relies on linear regression. In particular, we use a window of 50 estimated gaze angles to compute the regression parameters using 2 independent regression models, that is, one per gaze axis in 2D. The estimated parameters are used to predict the next 5 frames into the future. We apply our trained gaze estimation network on the test subset to estimate gaze for the first 50 frames of each sequence, compute linear regression parameters on them and predict the next 5 frames. Table \ref{tab:prediction} summarizes the obtained results. We can see that the error increases with time, which is expected. Furthermore, we can also observe that such a simple approach works fairly well for fixation and smooth pursuit sequences, which account for most of the dataset samples. However, the obtained error for saccades is large, as observed from p95 values. This is also expected mainly for two reasons. First, saccades usually follow a ballistic trajectory, thus not properly modeled with a linear model. Second, saccades have a duration of about 20 to 200ms; therefore, linear regression cannot predict extremely short saccades that happen after the initialization window.

\begin{table}[t!]
\centering
        \begin{tabular}{|c|c|c|c|c|c|}
            \hline
            Time step & Average & p50 & p75 & p95 \\
            \hline
            1 (10ms) & 5.28 & 4.56 & 6.73 & 11.89 \\
            2 (20ms) & 5.32 & 4.57 & 6.79 & 11.99 \\
            3 (30ms) & 5.37 & 4.61 & 6.83 & 12.13 \\
            4 (40ms) & 5.41 & 4.63 & 6.87 & 12.30 \\
            5 (50ms) & 5.46 & 4.65 & 6.92 & 12.48 \\
            \hline
        \end{tabular}
    \caption{Baseline performance for gaze prediction, reported in terms of angular error between predicted and ground truth 3D gaze vectors, in degrees.} 
    \label{tab:prediction}
\end{table}

\subsection{Semantic eye segmentation} \label{sec:seg_method}
In order to set a baseline for spatio-temporal eye-segmentation algorithms, we chose to train a deep-CNN on 1,605 images whose corresponding hand-annotated semantic segmentation masks are provided. The network follows an encoder-decoder architecture loosely based on SegNet \cite{SegNet}. Modifications include a power efficient version that contains 7 downscaler blocks for the encoder and 7 upscaler blocks for the decoder, with each block containing separable convolution that factorizes depth-wise convolution and a 1x1 convolution to optimize for computation cost. We use LeakyReLU activation and multiplicative skip connections with guide layers that reduce down the layers to a single channel to pass to the decoder, which again reduces computational cost.

We trained the network for 150 epochs with ADAM optimizer, with initial learning rate of 0.004 and batch size of 128. We utilized random rotation and intensity scaling for augmentations. Model training and inference was performed on downsampled images of size 128$\times$128 pixels. The total number of parameters of the model was 40k and the resulting model size was only 300KB.

The trained network is tested on the hidden set of test samples without exploiting any temporal information, achieving a mIoU score of 0.841, see Table.~\ref{tab:sem_baseline} for complete results.

\begin{table}[t!]
\label{tab:sem_baseline}
\centering
    \begin{tabular}{|c|c|c|c|c|}
        \hline
         Background & Sclera & Iris & Pupil & Average  \\
         \hline
         0.971 & 0.674 & 0.882 & 0.835 & 0.841 \\
         \hline
    \end{tabular}
    \caption{Baseline performance (mIOU) for sparse semantic segmentation without the use of temporal information.}
\end{table}

\section{Conclusion}
We have presented OpenEDS2020, a novel dataset of eye-image sequences consisting of up to 80 subjects of varied appearance performing different gaze-elicited tasks. The dataset consists of two subsets of data, one devoted to gaze prediction and another devoted to eye semantic segmentation, with the goal of fostering the exploitation of spatio-temporal information to improve the current state of the art on both tasks. Obtained results with proposed baseline methods demonstrate the usefulness of the data, and serve as a benchmark for future approaches coming from computer vision, machine learning and eye tracking fields. 

{\small
\bibliographystyle{ieee}
\bibliography{egbib}
}

\clearpage

\end{document}